\documentclass[runningheads]{llncs}
\usepackage[T1]{fontenc}
\usepackage{graphicx}
\usepackage{amsmath}
\usepackage{amssymb}
\usepackage{xcolor}
\usepackage{tabularx}
\usepackage{multirow}
\usepackage{multicol}
\usepackage{booktabs}
\usepackage{hyperref}
\usepackage{orcidlink}
\usepackage{makecell}
\usepackage[dvipsnames]{xcolor}
\begin{document}
\title{VENUS: Visual Editing with Noise Inversion Using Scene Graphs}
%
\titlerunning{VENUS}
\newcommand{\repeatthanks}{\textsuperscript{\thefootnote}}

%
\author{Thanh-Nhan Vo\orcidlink{0009-0007-8403-1240}\inst{1,2} 
\and
Trong-Thuan Nguyen\orcidlink{0000-0001-7729-2927}\inst{1,2}$^\spadesuit$ 
\and \\
Tam V. Nguyen\orcidlink{0000-0003-0236-7992}\inst{3} 
\and
Minh-Triet Tran\orcidlink{0000-0003-3046-3041}\inst{1,2}}

\authorrunning{Thanh-Nhan Vo et al.}

\institute{University of Science, VNU-HCM, Vietnam 
\and
Vietnam National University, Ho Chi Minh City, Vietnam
\and
University of Dayton, U.S.A.\\
\email{\{vtnhan,ntthuan\}@selab.hcmus.edu.vn,} \\ 
\email{tmtriet@fit.hcmus.edu.vn}, \email{tamnguyen@udayton.edu}}

\def\thefootnote{$^\spadesuit$}\footnotetext{Corresponding Author}\def\thefootnote{\arabic{footnote}}
\maketitle
\begin{abstract}
State-of-the-art text-based image editing models often struggle to balance background preservation with semantic consistency, frequently resulting either in the synthesis of entirely new images or in outputs that fail to realize the intended edits. In contrast, scene graph-based image editing addresses this limitation by providing a structured representation of semantic entities and their relations, thereby offering improved controllability. However, existing scene graph editing methods typically depend on model fine-tuning, which incurs high computational cost and limits scalability. To this end, we introduce \textbf{VENUS} (\textit{\textbf{V}}isual \textit{\textbf{E}}diting with \textit{\textbf{N}}oise inversion \textit{\textbf{U}}sing \textit{\textbf{S}}cene graphs), a training-free framework for scene graph-guided image editing. Specifically, VENUS employs a split prompt conditioning strategy that disentangles the target object of the edit from its background context, while simultaneously leveraging noise inversion to preserve fidelity in unedited regions. Moreover, our proposed approach integrates scene graphs extracted from multimodal large language models with diffusion backbones, without requiring any additional training. Empirically, VENUS substantially improves both background preservation and semantic alignment on PIE-Bench, increasing PSNR from 22.45 to 24.80, SSIM from 0.79 to 0.84, and reducing LPIPS from 0.100 to 0.070 relative to the state-of-the-art scene graph editing model (SGEdit). In addition, VENUS enhances semantic consistency as measured by CLIP similarity (24.97 vs. 24.19). On EditVal, VENUS achieves the highest fidelity with a 0.87 DINO score and, crucially, reduces per-image runtime from 6–10 minutes to only 20–30 seconds. Beyond scene graph-based editing, VENUS also surpasses strong text-based editing baselines such as LEDIT++ and P2P+DirInv, thereby demonstrating consistent improvements across both paradigms.


\end{abstract}

\section{Introduction}\label{sec:intro}
Editing visual scenes in a controllable and efficient manner remains a fundamental challenge in computer vision. Broadly, two paradigms have been explored: text-driven diffusion editing and scene graph-based editing. On the one hand, text-only methods~\cite{brack2024ledits++,ju2023direct} offer a convenient and intuitive interface, making them highly accessible and flexible across diverse editing scenarios. On the other hand, scene graph-based approaches~\cite{zhang2024sgedit} provide more precise control over object relations and spatial layout, underscoring the importance of structured scene representations for reliable visual editing. However, despite these advantages, text-driven methods often suffer from semantic imprecision, leading to object misplacement and background drift. Conversely, scene graph-based methods typically require costly training or inference-time fine-tuning, resulting in long editing latencies (6–10 minutes per image). Therefore, the fundamental trade-off between semantic consistency and background fidelity remains unresolved.


\noindent\textbf{Contributions of this Work.} We propose \textbf{VENUS} (\textit{\textbf{V}}isual \textit{\textbf{E}}diting with \textit{\textbf{N}}oise inversion \textit{\textbf{U}}sing \textit{\textbf{S}}cene Graphs), a training-free framework for scene graph-guided image editing. VENUS introduces a split prompt conditioning strategy that disentangles the target object to be edited from the preserved background, thereby enabling diffusion models to perform controllable semantic editing without compromising background fidelity. Furthermore, by leveraging multimodal large language models (MLLMs) for automatic scene graph extraction and refinement, VENUS integrates seamlessly with existing diffusion backbones and significantly reduces editing latency. Notably, without requiring any additional training or inference-time fine-tuning, VENUS outperforms SGEdit~\cite{zhang2024sgedit} on the PIE-Bench dataset~\cite{ju2023direct}, reducing LPIPS from 0.100 to 0.070, increasing PSNR from 22.45 to 24.80, improving SSIM from 0.790 to 0.837, and raising the CLIP from 24.19 to 24.97. Beyond outperforming scene graph-based methods, VENUS also consistently surpasses strong text-based baselines, including LEDIT++\cite{brack2024ledits++} and P2P+DirInv~\cite{ju2023direct}, demonstrating its effectiveness across editing paradigms.

\section{Related Work}\label{sec:related}
\vspace{-0.5em}
\subsection{Scene Graph to Image Synthesis and Editing}\label{subsec:sg}
\vspace{-0.5em}
Scene graphs represent visual scenes as nodes (objects) and edges (spatial or semantic relations), which have been successfully applied to image captioning~\cite{Nguyen_2021_ICCV}, cross-modal retrieval~\cite{yoon2021image}, image generation~\cite{johnson2018image}, and image editing~\cite{zhang2024sgedit}. For visual synthesis, early work such as SG2IM~\cite{johnson2018image} employed a two-stage layout refinement pipeline, whereas SATURN~\cite{vo2025saturn} transformed scene graphs into structured textual prompts, thereby leveraging the capabilities of modern text-to-image diffusion models without requiring additional training. For image editing, SGEdit~\cite{zhang2024sgedit} integrates scene graph parsing, concept grounding, and large language model–guided prompting into diffusion models, achieving strong accuracy and fidelity. Taken together, these works highlight the role of scene graphs as a structured and interpretable interface that is particularly well-suited for both generation and editing within contemporary diffusion-based frameworks.

\subsection{LLM-based Image Synthesis}\label{subsec:llm}
In recent years, LLM-based methods have emerged as highly effective solutions. By leveraging the multimodal reasoning capabilities of LLMs across text, visual inputs, and structured representations such as scene graphs, many methods~\cite{sun2024autoregressive,qin2024diffusiongpt} have achieved remarkable performance in producing high-quality, semantically aligned images. Unlike traditional pipelines that rely solely on text encoders, LLMs can serve as high-level planners, determining scene layouts and object compositions prior to the generation stage. For example,~\cite{zhang2024sgedit,wang2024genartist} employ LLMs to generate structured layouts or scene graphs, which in turn guide diffusion or autoregressive models, yielding more controllable and faithful synthesis.

\subsection{Diffusion-Based Image Editing}\label{subsec:image_editing}
DDPM~\cite{ho2020denoising} and Stable Diffusion~\cite{rombach2022high} have greatly enhanced the ability to synthesize diverse, photorealistic images with high detail and naturalness. As a result, diffusion-based models have been widely adopted across domains, including medicine, the arts, and media.
Meanwhile, image editing is particularly notable, as it demands a fine-grained understanding of both natural language and visual semantics. Existing methods can be broadly categorized into three main types. The first category consists of training-based methods, such as DiffusionCLIP~\cite{kim2022diffusionclip} and InstructPix2Pix~\cite{brooks2023instructpix2pix}, which train models from scratch on large-scale, language-guided datasets. The second category includes test-time fine-tuning methods, such as StyleDiffusion~\cite{wang2023stylediffusion} and Imagic~\cite{kawar2023imagic}, which adapt pretrained models during inference to maximize representational capacity. Finally, there are training-free and fine-tuning-free methods, including LEDITS++~\cite{brack2024ledits++}, Plug-and-Play (PnP)~\cite{tumanyan2023plug}, Prompt-to-Prompt (P2P)~\cite{Hertz2022PrompttoPromptIE}, and MasaCtrl~\cite{cao2023masactrl}, which operate on pretrained models, typically by modifying latent representations.

\subsection{Discussion}\label{subsec:discuss}

\noindent\textbf{Limitations of Previous Methods.}
As discussed in Secs.\ref{subsec:sg}, \ref{subsec:llm}, and \ref{subsec:image_editing}, existing methods face significant challenges in jointly preserving background fidelity and maintaining semantic consistency during editing. On the one hand, text-driven diffusion models often introduce background drift or semantic misalignment, producing outputs that deviate from the intended edits. On the other hand, scene graph-based methods, while offering greater controllability, typically require additional training or fine-tuning at inference time\cite{zhang2024sgedit,kawar2023imagic}. This requirement increases computational and reproduction cost and leads to substantial editing latency, with processing times commonly ranging from six to ten minutes per image. Specifically, these limitations severely restrict scalability and hinder practicality in real-world scenarios. Consequently, the \textit{trade-off between semantic alignment and background preservation} remains unresolved in prior work, motivating the need for a more efficient and training-free solution.

\noindent\textbf{Advantages of Our Approach.} VENUS addresses the aforementioned limitations by \textit{leveraging scene graphs as structural priors while preserving the accessibility of text-based editing}. In particular, our approach integrates scene graph guidance directly into the diffusion process through a training-free noise inversion mechanism, which ensures that edits are accurately localized while leaving unedited regions unaffected. This design \textit{retains the fine-grained relational control afforded by scene graphs}, while simultaneously benefiting from the high visual fidelity of state-of-the-art diffusion models. Furthermore, VENUS eliminates the need for additional training or inference-time fine-tuning, thereby \textit{reducing editing latency} from several minutes to only a few seconds. Consequently, our approach enables controllable, efficient, and high-quality scene graph–based image editing, while remaining fully compatible with existing diffusion backbones.

\section{Problem Formulation}\label{subsec:formulation}

In this work, we address the problem of editing images from scene graphs, where a scene graph is defined as a set of triplets, each augmented with bounding boxes and object categories. Formally, we define the scene graph as in Eqn~\eqref{definegraph}. 
\begin{equation}
    \mathcal{G} = \{(s_i, r_i, o_i) \}^{N}_{i=1}, \quad N = |\mathcal{G}|
    \label{definegraph}
\end{equation}
where $s_i, r_i$ and $o_i$ denote the subject, relation, and object of the $i$-th triplet.

Given a source image $\mathcal{I}$, our approach first detects objects and relations to construct an initial scene graph $\mathcal{G}$. The user then interactively modifies $\mathcal{G}$ (by selecting objects, editing attributes, or altering relations) to obtain a target graph $\mathcal{G}^*$. We mathematically express the interactive editing task in Eqn~\eqref{definetask}. 
\begin{equation}
    (\mathcal{I}, \Delta \mathcal{G}) \xrightarrow{f_\theta} \hat{\mathcal{I}}
    \label{definetask}
\end{equation}
where $\Delta \mathcal{G} = \mathcal{G}^* - \mathcal{G}$ represents the user-specified changes, and $\theta$ denotes the parameters of the model. The editor generates the edited image $\hat{I}$ by applying the modifications in $\Delta \mathcal{G}$ to $\mathcal{I}$, while preserving visual fidelity in unchanged regions.

\section{Our Proposed Approach}\label{sec:approach}
\begin{figure}[!htbp]
    \centering
    \includegraphics[width=.85\linewidth]{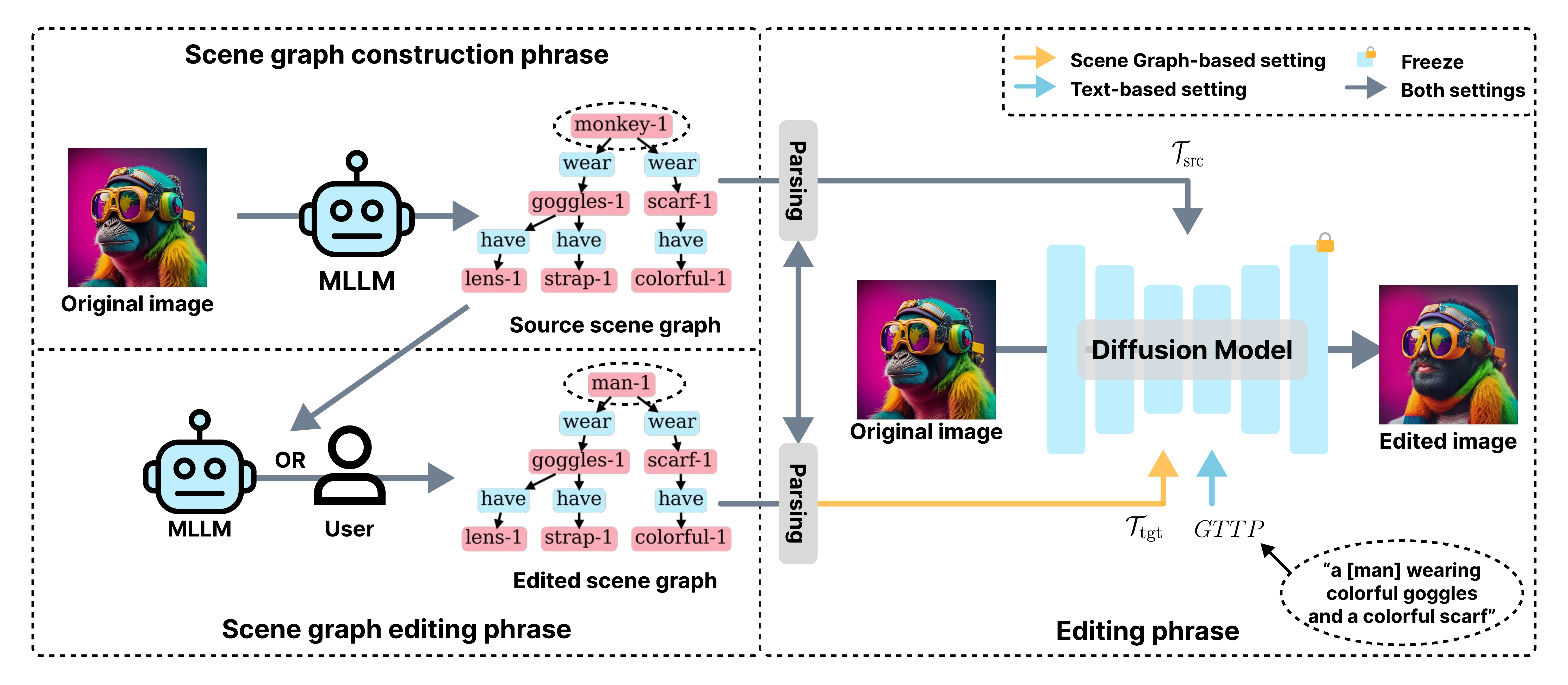}
    \vspace{-\baselineskip}
    \caption{An illustration of our VENUS approach. A scene graph is constructed from the input image, then edited by either the MLLM or the user. The edited graph is parsed into structured prompts, which condition a frozen diffusion model to edit the image. \textbf{(Best viewed in color and with zoom.)}}
    \label{fig:method}
\end{figure}

Fig.~\ref{fig:method} presents our proposed approach, termed VENUS, which leverages multimodal LLM to extract and refine scene graphs from an image. We then parse the graph, prune irrelevant relations, and preserve key ones, using the result as a conditioning signal for precise and controllable diffusion-based image editing.

\subsection{Scene Graph Construction and Refinement via MLLM} \label{sgconstruction}
To leverage the spatial understanding capabilities of scene graphs, we employ Qwen-VL-2.5-7B, an advanced lightweight MLLM, to extract the scene graph $\mathcal{G}$ from an input image $\mathcal{I}$. Typically, after parsing the scene graph with an MLLM, a user may manually manipulate it. Alternatively, the MLLM itself can be employed to perform automated edits. In the automatic scenario, we generate an edited scene graph $\mathcal{G}’$ conditioned on an instruction $\mathcal{T}$, as defined in Eqn.~\eqref{eqn:1}.
\begin{equation}
    \mathcal{G} = \text{MLLM}(\mathcal{I}), \quad \mathcal{G}' = \text{MLLM}(\mathcal{I}, \mathcal{G}, \mathcal{T})
    \label{eqn:1}
\end{equation}

In the Eqn.~\eqref{eqn:1}, the two-step process ensures that the original scene graph $\mathcal{G}$ and the edited scene graph $\mathcal{G}’$ remain structurally aligned. Thereby, this process preserves objects, subjects, and unaffected relations while simultaneously capturing the semantic modifications specified by the editing instruction.

\subsection{Target prompt and source prompt construction}
\subsubsection{Target Prompt Construction.} We first parse scene graphs into textual descriptions for text-to-image models. Unlike the previous method~\cite{vo2025saturn}, which focuses on image generation, our goal is editing, thus we construct two separate prompts: one for edited content $\mathcal{T}_{\mathcal{G}_\text{new}}$ and one for preserved background $\mathcal{T}_{\mathcal{G}_\text{bgd}}$. We adopt a similar strategy by converting each relation triplet into a concise textual phrase of the form $t = (s, r, o)$ where $s$, $r$, and $o$ denote the subject, relation, and object, respectively. These phrases are then concatenated to form a raw caption, which is defined as $\mathcal{T}_{\mathcal{G}} = [t_1, t_2, \dots, t_N], \quad t_i = (s_i, r_i, o_i) \in  \mathcal{G}$
For example, the triplet (``dog'', ``sitting on'', ``bench'') becomes the phrase ``dog sitting on bench''. To better guide the image generation process, we further decompose the target caption $\mathcal{T}_{\text{tgt}}$ into two components: $\mathcal{T}_{\mathcal{G}_{\text{new}}}$ and $\mathcal{T}_{\mathcal{G}_{\text{bgd}}}$, defined in Eqn.~\eqref{eqn:3}.
\begin{equation}
    \mathcal{T}_{\text{tgt}} = \mathcal{T}_{\mathcal{G}_{\text{new}}} + \mathcal{T}_{\mathcal{G}_{\text{bgd}}}
    \label{eqn:3}
\end{equation}
where $\mathcal{T}_{\mathcal{G}_\text{new}}$ denotes the set of relation phrases that appear exclusively in $\mathcal{G}'$ but not in $\mathcal{G}$, and $\mathcal{T}_{\mathcal{G}_\text{bgd}}$ represents the set of relations shared by both $\mathcal{G}$ and $\mathcal{G}'$. The corresponding subgraphs $\mathcal{G}_{\text{new}}$ and $\mathcal{G}_{\text{bgd}}$ are formally defined in Eqn.~\eqref{eqn:4}.

\begin{equation}
    \mathcal{G}_{\text{new}} = \mathcal{G}' \setminus \mathcal{G}, \quad 
    \mathcal{G}_{\text{bgd}} = \mathcal{G} \cap \mathcal{G}'
    \label{eqn:4}
\end{equation}

In the Eqn.~\eqref{eqn:4}, the separation emphasizes novel or edited content during image synthesis and mitigates potential information loss caused by token truncation (e.g., exceeding the 77-token limit of CLIP's text encoder~\cite{radford2021learning}).

\subsubsection{Source Prompt Construction}
Our approach supports the diffusion model in inverting image noise while preserving the vanilla scene structure, we do not utilize the full scene graph $\mathcal{G}$. Instead, we selectively retain only the subset of relation triplets that are common to both the vanilla and edited graphs, i.e., $\mathcal{G} \cap \mathcal{G}'$. These relations form the background caption, denoted as $\mathcal{T}_{\mathcal{G}_{\text{bgd}}}$. Accordingly, the source caption is defined as $\mathcal{T}_{\text{src}} = \mathcal{T}_{\mathcal{G}_{\text{bgd}}}$.
This design choice ensures that the source prompt encodes only the stable semantic components, thereby facilitating faithful reconstruction of the input image during inversion.

\subsection{Image Editing}
Our approach can integrate with various diffusion-based editing backbones. In this work, we choose LEDIT++~\cite{brack2024ledits++} due to its strong performance in semantic editing and background preservation (see Table~\ref{tab:compare}). In addition, our approach enables unified conditioning on both scene graphs and textual inputs.
\subsubsection{Scene graph-based editing}
In our approach, we leverage the LEDIT++~\cite{brack2024ledits++} framework for scene graph-based image editing, in which the source $\mathcal{T}_{\text{src}}$ and target prompts $\mathcal{T}_{\text{tgt}}$, are used to guide the diffusion model throughout both the inversion and editing phases. Unlike vanilla LEDIT++ that utilize editing prompts in the form of short attribute-level instructions (e.g., ``+glasses'', ``-hat''), we intentionally omit such simplistic expressions. Instead, we incorporate structurally rich prompts derived from scene graphs to provide more comprehensive semantic control over the editing process, as formally defined in Eqn.~\eqref{eqn:sgbased}.
\begin{equation}
    \label{eqn:sgbased}
    \mathcal{I_{\text{edited}}} = \mathcal{M}(\mathcal{T}_{\text{src}}, \mathcal{T}_{\text{tgt}})
\end{equation}
where $\mathcal{M}$ denotes a text-guided image editing model, and $\mathcal{I}_{\text{edited}}$ is the output image after applying semantic edits guided by both source and target prompts.

\subsubsection{Text-based editing}
For text-based image editing, we also use LEDIT++ as the backbone of our framework. We observe that existing benchmarks predominantly evaluate edited images based on their CLIP similarity with the target prompt. To align with this evaluation, we utilize the ground-truth target prompt (GTTP) as input during editing. In addition, we incorporate the source prompt $\mathcal{T}_{\text{src}}$, which is constructed from the original scene graph, as auxiliary guidance during inversion and generation, as formally defined in Eqn.~\eqref{eqn:textbased}.
\begin{equation}
    \label{eqn:textbased}
    \mathcal{I_{\text{edited}}} = \mathcal{M}(\mathcal{T}_{\text{src}}, \text{GTTP})
\end{equation}
where $\mathcal{M}$ denotes a text-guided image editing model, and $\mathcal{I}_{\text{edited}}$ is the output image after applying semantic edits guided by both source and target prompts.

\subsubsection{Classifier-Free Guidance} 
Prompts are injected during both inversion and editing through classifier-free guidance (CFG)~\cite{Ho2022ClassifierFreeDG}, ensuring that $T_{\text{src}}$ and $T_{\text{tgt}}$ continuously influence the internal representations and effectively guide the image generation process. Specifically, this mechanism leads to edits that are more consistent and of higher visual quality. The inference-time CFG is defined as $\epsilon_{\text{pred}} = \epsilon_{\text{null text}} + s (\epsilon_{\text{text}} - \epsilon_{\text{null text}})$, where $\epsilon_{\text{pred}}$ denotes the final predicted noise, $\epsilon_{\text{text}}$ the prediction conditioned on the prompt, $\epsilon_{\text{null text}}$ the unconditioned prediction, and $s > 1$ the guidance scale controlling the strength of conditioning.

\section{Experiment Results}\label{sec:experiment}
\begin{table}[!t]
\centering
\renewcommand{\arraystretch}{1.1}
\caption{Comparison of different models on background preservation and semantic consistency. \textbf{GTTP} denotes the \textit{Ground-Truth Target Prompt} used during editing and evaluation. Best score in \textbf{bold}, second best in \underline{underline}.}
\vspace{-2mm}
\begin{tabular}{l|c|ccc|c}
\toprule
\textbf{Method} & \textbf{Backbone} & \textbf{PSNR $\uparrow$} & \textbf{SSIM $\uparrow$} & \textbf{LPIPS $\downarrow$} & \makecell[c]{\textbf{CLIP} \\ \textbf{Whole} $\uparrow$} \\

\midrule
\multicolumn{6}{l}{\textit{\textbf{Text-based Editing Methods}}} \\
P2P~\cite{Hertz2022PrompttoPromptIE}  &--& 17.87 & 0.7114 & 0.2088 & 25.01 \\
Pix2Pix-Zero~\cite{brooks2023instructpix2pix}    &--& 20.44 & 0.7467 & 0.1722 & 22.80 \\
MasaCtrl~\cite{cao2023masactrl}        &--& 22.17 & 0.7967 & 0.1066 & 23.96 \\
PnP~\cite{tumanyan2023plug}             &--& 22.28 & 0.7905 & 0.1134 & \underline{25.41} \\
PnP + DirInv~\cite{ju2023direct}     &--& 22.46 & 0.7968 & 0.1061 & \underline{25.41} \\
P2P + DirInv~\cite{ju2023direct} &--& \textbf{27.22} & \textbf{0.8476}  & \textbf{0.0545} & 25.02\\
LEDIT++$_{\text{(only GTTP)}}$~\cite{brack2024ledits++} &SDv2.1& \underline{23.18} & \underline{0.8218} & \underline{0.0855} & \textbf{26.85}\\
\midrule
VENUS w$/_{\text{GTTP}}$&LEDIT++& \underline{23.54}
& \underline{0.8288} 
& \underline{0.084}
& \textbf{26.89} \\
VENUS w$/_{\text{GTTP}}$&P2P+DirInv&\textbf{27.25}&\textbf{0.853}&\textbf{0.05}&\underline{25.99}\\
\midrule
\multicolumn{6}{l}{\textit{\textbf{Scene graph-based Editing Methods}}} \\

DiffSG~\cite{yang2022diffusion}          &--& 9.35   & 0.41  & 0.55 & 12.45 \\
SIMSG~\cite{dhamo2020semantic}           &--& \underline{19.48}  & \underline{0.70}  & \underline{0.40} & \underline{20.40} \\
SGEdit~\cite{zhang2024sgedit}          &SDv2.1& \textbf{22.45}  & \textbf{0.79} & \textbf{0.10} & \textbf{24.19} \\
\midrule
VENUS (Ours) &LEDIT++& \textbf{24.80} & \textbf{0.837  } & \textbf{0.070} & \textbf{24.97}\\
\bottomrule
\end{tabular}
\label{tab:compare}
\end{table}
\subsection{Implementation Details}\label{subsec:details}
\subsubsection{Model Configuration.} 
Our experiments build upon the LEDIT++ framework~\cite{brack2024ledits++}, employing Stable Diffusion v2.1 (SDv2.1)~\cite{rombach2022high} as the generative backbone to enhance both visual fidelity and editability. Unless otherwise stated, we set the number of sampling steps to 50, the skip parameter to 25, and the random seed to 42. For semantic understanding and editing guidance, we use Qwen-VL 2.5 (7B)~\cite{qwen2.5-VL} as the multimodal language model to extract scene graphs from images and suggest structural edits. To prevent overly long prompts and token truncation in the text encoder, the number of extracted relations is capped at 15. All experiments are conducted on the NVIDIA A100 GPU with 80GB.
\subsubsection{Dataset and Metric.}
We evaluate on PIE-Bench~\cite{ju2023direct}, a text-based benchmark designed for diverse image editing tasks. In addition, we report performance using PSNR~\cite{PSNR}, SSIM~\cite{1284395}, and LPIPS~\cite{zhang2018perceptual} for background preservation, and CLIP similarity~\cite{radford2021learning} for semantic alignment with the target prompt.

\subsection{Quantitative Results}\label{subsec:quant}

Table~\ref{tab:compare} reports a comparison between our approach and state-of-the-art baselines across four key metrics.
Compared to the strongest scene graph-based model (SGEdit), our approach improves PSNR from 22.45 to 24.80 and SSIM from 0.79 to 0.84, while reducing LPIPS from 0.100 to 0.070. The CLIP similarity also increases from 24.19 to 24.97, indicating improved semantic alignment without compromising content fidelity. In the text-based setting, VENUS surpasses strong baselines, such as P2P+DirInv and LEDIT++, by a significant margin. With the LEDIT++ backbone under the GTTP configuration, we obtain a CLIP score of 26.89 (+0.04 over LEDIT++), along with measurable improvements in background preservation. Using the P2P+DirInv backbone, our approach yields a +0.97 increase in the CLIP score and further improves preservation metrics. This demonstrates that our split-prompt design forces the model to respect edited relations while avoiding catastrophic drift in background context, a balance that previous direct-prompting approaches fail to achieve. We emphasize that conditioning on the GTTP in PIE-Bench is not an unfair bias but mirrors the standard operating mode of conventional text-based image editing, where the user-provided target prompt directly guides the edit. Our text-based setting therefore faithfully simulates real-world usage, ensuring a fair, apples-to-apples comparison with existing text-based baselines (e.g., LEDIT++, P2P+DirInv, SGEdit). This result highlights that VENUS is not restricted to scene graph editing: even in the purely text-based setting, VENUS achieves the highest CLIP similarity (26.89) and improved fidelity, surpassing previous models.
\begin{table}[!t]
\centering
\caption{Comparison on EditVal. Accuracy is measured using OwL-ViT~\cite{minderer2022simple}, and fidelity is measured using DINO~\cite{caron2021emerging} (higher is better). Reported runtime corresponds to the average editing time per image. Best score in \textbf{bold}, second best in \underline{underline}.}
\vspace{-2mm}
\begin{tabular}{l|cc|c}
\toprule
\textbf{Method} & 
\makecell[c]{\textbf{Accuracy} \\ \Xhline{0.6pt} \textbf{OwL-ViT} $\uparrow$} & 
\makecell[c]{\textbf{Fidelity} \\ \Xhline{0.6pt} \textbf{DINO} $\uparrow$}&
\makecell[c]{\textbf{Time}\\ \textbf{per image}}\\
\midrule
\multicolumn{4}{l}{\textit{\textbf{Scene graph-based Editing Methods}}}  \\
SIMSG         & 0.11 & 0.57&--\\
DiffSG        & 0.01 & 0.13&--\\
SGEdit & \textbf{0.53} & \underline{0.83}&\underline{6-10m} \\
\midrule
VENUS (Ours) & \underline{0.32} & \textbf{0.87}&\textbf{20-30s}\\

\bottomrule
\end{tabular}
\label{tab:EditVal}
\end{table}

Additionally, we evaluate our approach on the EditVal benchmark using OwL-ViT~\cite{minderer2022simple} and DINO~\cite{caron2021emerging} as automatic metrics in Table~\ref{tab:EditVal}. Our approach attains an OwL-ViT accuracy of 32\% and a DINO fidelity score of 87\%. While the fidelity surpasses all baselines, the accuracy falls short of SGEdit (53\%). We attribute this gap to two primary factors: (1) EditVal is built on MS-COCO~\cite{lin2014microsoft}, which features diverse aspect ratios and resolutions, whereas our pipeline is based on Stable Diffusion v2.1~\cite{rombach2022high} without specific adaptation for such variations; and (2) EditVal emphasizes precise spatial localization of edits, a criterion not explicitly optimized by LEDIT++ backbones. We note that this limitation is common across many diffusion-based models. Moreover, \textit{without any fine-tuning phase, VENUS significantly reduces editing time to just 20–30 seconds per image}.

\begin{table}[!t]
\centering
\renewcommand{\arraystretch}{1.4}
\caption{Comparison of editing performance with constructed and ground-truth target prompts. Scores are ImageReward (higher is better) and FID (lower is better), evaluated on PIE-Bench. Best score in \textbf{bold}, second best in \underline{underline}.}
\vspace{-2mm}
\begin{tabular}{l|c|cc|c}
\toprule
\multirow{2}{*}{\textbf{Method}} & \multirow{2}{*}{\textbf{Backbone}} & \multicolumn{2}{c|}{\textbf{ImageReward $\uparrow$}} & \multirow{2}{*}{\textbf{FID $\downarrow$}} \\
\cmidrule(lr){3-4}
& & \makecell{Constructed} & \makecell{GT} & \\
\midrule
LEDIT++ (only GTTP) & SDv2.1 & \underline{0.514} & \textbf{0.636} & \underline{53.01} \\
VENUS w$/_{\text{GTTP}}$ & LEDIT++ & 0.363 & \underline{0.624} & 53.54 \\
VENUS (Ours) & LEDIT++ & \textbf{0.716} & 0.071 & \textbf{47.38} \\
\bottomrule
\end{tabular}
\label{tab:ImageReward}
\end{table}

\begin{figure}[!b]
  \centering
  \includegraphics[width=0.8\linewidth]{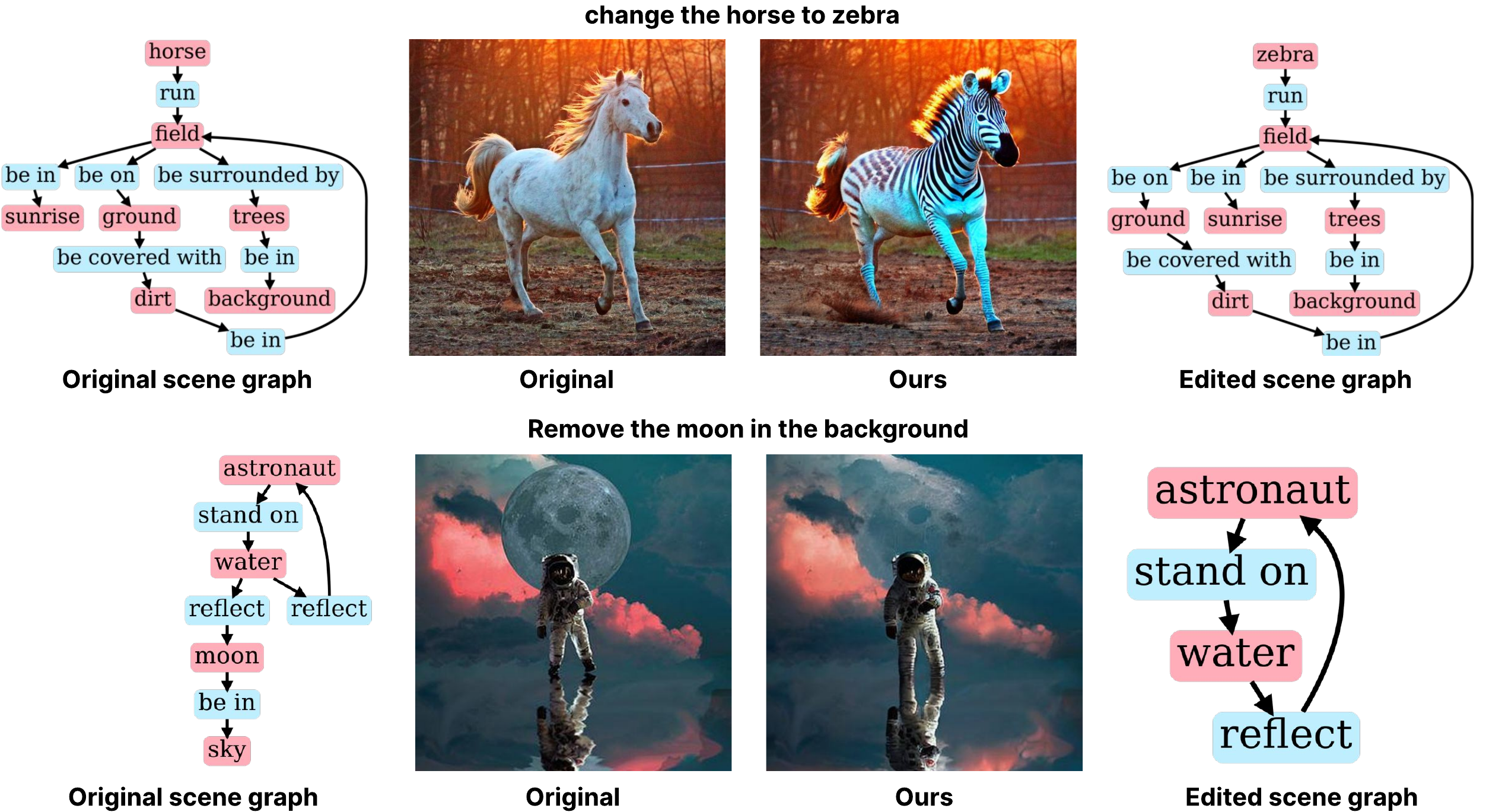}
  \vspace{-\baselineskip}
  \caption{Examples of scene graph guided image editing. Top row: changing the horse into a zebra by updating the corresponding node in the scene graph. Bottom row: removing the moon in the background by deleting its associated nodes and relations. \textbf{(Best viewed in color and with zoom.)}
  }
  \label{fig:sample1}
\end{figure} 
Table~\ref{tab:ImageReward} compares VENUS with text-based editing baselines using the ImageReward and FID metrics. With our constructed target prompts, the scene graph-based setting attains the highest ImageReward score of 0.716, outperforming both the text-based variant (0.363) and the LEDIT++ baseline with GTTP only (0.514).
When evaluated using ground-truth prompts, LEDIT++ (GTTP only) achieves the highest score (0.636), with our text-based variant following closely at 0.624. In terms of FID, our setting achieves the best visual quality (47.38), outperforming LEDIT++ (53.01) and our text-based variant (53.54).
\begin{figure}[!t]
  \centering
  \includegraphics[width=0.8\linewidth]{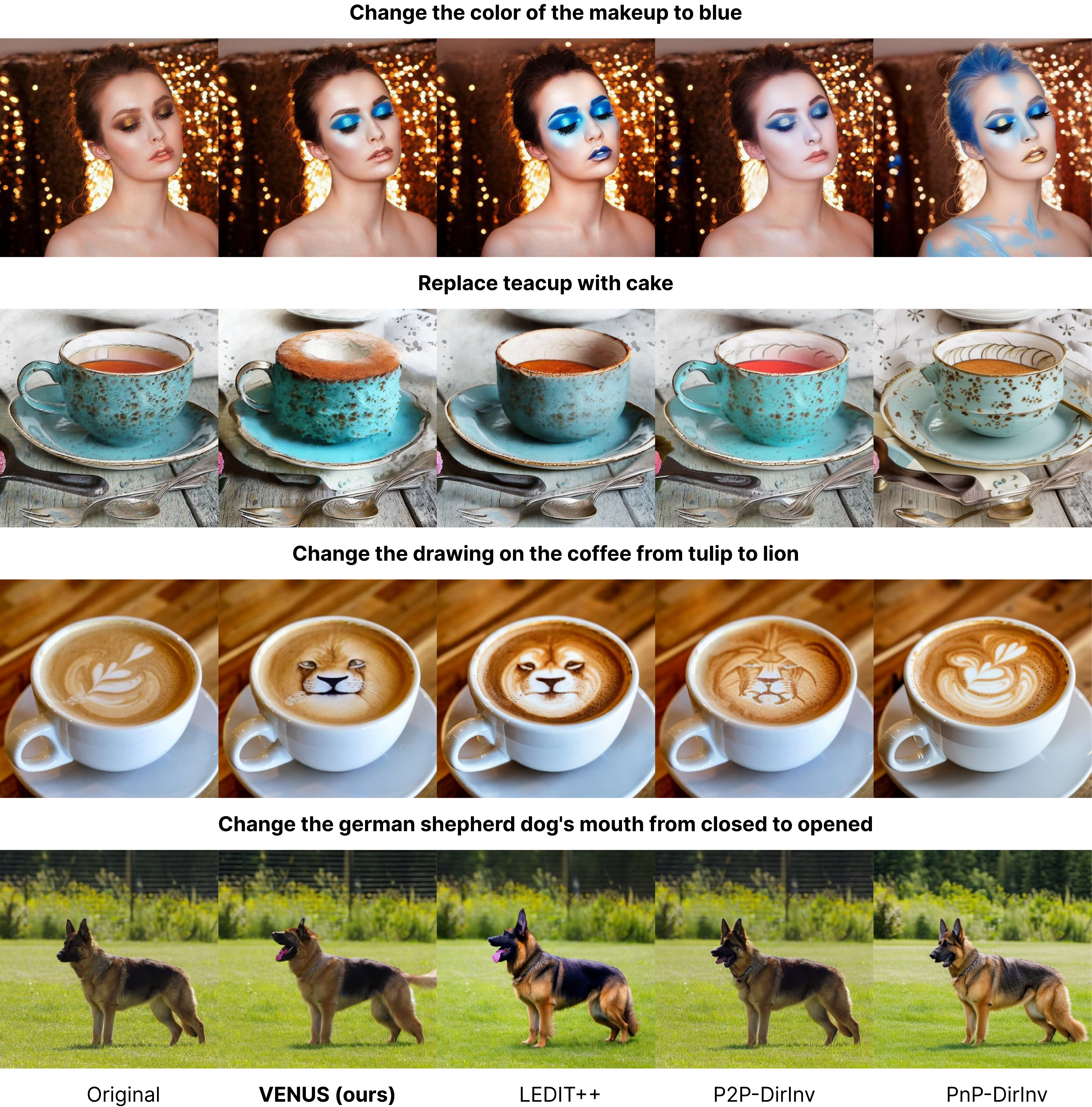}
  \vspace{-\baselineskip}
  \caption{Comparison of editing results across a variety of tasks. Compared to LEDIT++, P2P-DirInv, and PnP-DirInv, VENUS (ours) produces edits that are both semantically accurate and visually consistent. \textbf{(Best viewed in color and with zoom.)}}
  \label{fig:compare}
\end{figure}
\subsection{Qualitative Results}\label{subsec:quanl}

To demonstrate the effect of incorporating scene graph inputs into the generated outputs, Fig.~\ref{fig:sample1} showcases VENUS's ability to edit images based on complex scene graph modifications. The edited results exhibit a strong correspondence between the manipulated scene graphs and the visual outputs. With the proposed split prompt strategy, our approach can effectively handle structural edits, such as object removal. For example, in the second case, the moon is successfully removed from the background while preserving other scene elements.


In addition, we conduct a qualitative assessment to visually analyze the strengths and limitations of our approach. As illustrated in Fig.~\ref{fig:compare}, VENUS produces edited images with superior background preservation and semantic consistency compared to state-of-the-art text-based editing models~\cite{brack2024ledits++,ju2023direct}. The proposed approach more accurately focuses on the intended editing regions while minimizing unintended changes, thereby enhancing background fidelity.

\noindent


\subsection{Ablations Study}
\begin{table}[!t]
\centering
\renewcommand{\arraystretch}{1.1}
\caption{Evaluating the contribution of the Src prompt to background preservation and semantic consistency. w/o$_{\text{ Src prompt}}$ removes the source prompt during inversion; VENUS w/$_{\text{GTTP}}$ replaces our constructed target prompt with the ground-truth target prompt from PIE-Bench. Best score in \textbf{bold}, second best in \underline{underline}.}
\vspace{-2mm}
\begin{tabular}{l|ccc|c}
\toprule
\textbf{Method} & \textbf{PSNR $\uparrow$} & \textbf{SSIM $\uparrow$} & \textbf{LPIPS $\downarrow$} & 
 \makecell[c]{\textbf{CLIP} \\ \textbf{Whole} \(\uparrow\)} \\
\midrule
\multicolumn{5}{l}{\textit{\textbf{Scene Graph-based Editing Approaches}}} \\
\textbf{VENUS} 
& \textbf{24.80} 
& \textbf{0.837}
& \textbf{0.070}
& \textbf{24.97} \\

w/o$_{\text{ Src prompt}}$
& \underline{23.42}
& \underline{0.820}
& \underline{0.082}
& \underline{24.96}\\
\midrule
\multicolumn{5}{l}{\textit{\textbf{Text-based Editing Approaches}}} \\
\textbf{VENUS} w/$_{\text{GTTP}}$
& \textbf{23.54}
& \textbf{0.829}
& \textbf{0.084}
& \textbf{26.89}\\
LEDIT++~\cite{brack2024ledits++}& \underline{23.18} & \underline{0.822} & \underline{0.086} & \underline{26.85}\\
\bottomrule
\end{tabular}
\label{ablation}
\end{table}
\noindent\textbf{Enhancing Text-based Image Editing Models}
Table~\ref{tab:compareBackbone} reports that the effect of scene graph–based split prompt conditioning consistently improves text-based image editing across backbones (PnP+DirInv, P2P+DirInv, LEDIT++). The largest gain is observed in P2P+DirInv (PSNR 27.25, SSIM 0.853, LPIPS 0.05, +0.97 CLIP), indicating stronger semantic alignment with high fidelity. LEDIT++ achieves modest but improvements in fidelity and context retention.
\noindent\textbf{Effectiveness and Efficiency.} As shown in Table~\ref{ablation}, when the source prompt is removed, the performance drops significantly in terms of scene integrity, with a slight decrease in semantic consistency in both settings. Specifically, this indicates that the source prompt effectively improves background fidelity. Moreover, scene graph-based methods better maintain background details than text-based methods, highlighting the value of structured representations for coherent edits.

\noindent\textbf{Effect of Different MLLM Backbones.}
We evaluate our split-prompt strategy in both text-based and scene-graph-based settings using Qwen2.5-VL-instruct (7B), Phi-3.5-vision-instruct (4.15B), and InternVL2.5 (4B). As shown in Table~\ref{tab:compareMLLM}, performance is stable across backbones, with larger models (Qwen2.5-VL) tending to yield higher CLIP, while smaller models (InternVL2.5) excel in background preservation. These results indicate that our approach is robust to the choice of MLLM and allows trade-offs between semantic alignment and fidelity.

\begin{table}[!t]
\centering
\renewcommand{\arraystretch}{1}
\caption{Comparison of different backbones with and without our text-based setting approach for background preservation and semantic consistency.}
\vspace{-2mm}
\begin{tabular}{l|c|ccc|c}
\toprule
\textbf{Method} & \textbf{Backbone} 
& \textbf{PSNR} \(\uparrow\) 
& \textbf{SSIM} \(\uparrow\) 
& \textbf{LPIPS} \(\downarrow\) 
& \makecell[c]{\textbf{CLIP} \\ \textbf{Whole} \(\uparrow\)} \\
\midrule
\multicolumn{6}{l}{\textit{\textbf{Text-based Editing Methods}}} \\
PnP + DirInv~\cite{ju2023direct} & -- 
& 22.46 & 0.7968 & 0.1061 & \underline{25.41} \\
P2P + DirInv~\cite{ju2023direct} & -- 
& \textbf{27.22} & \textbf{0.8476} & \textbf{0.0545} & 25.02 \\
LEDIT++$_{\text{(only GTTP)}}$ & --
& \underline{23.18} & \underline{0.8218} & \underline{0.0855} & \textbf{26.85} \\
\midrule
\multicolumn{6}{l}{\textit{\textbf{Our Approach}}} \\
VENUS w$/_{\text{GTTP}}$ & PnP+DirInv 
& \makecell{22.35 \\ \scriptsize(-0.11)} 
& \makecell{0.8019 \\ \scriptsize(+0.0051)} 
& \makecell{0.1066 \\ \scriptsize(+0.0005)} 
& \makecell{\underline{26.21} \\ \scriptsize(+0.80)} \\

VENUS w$/_{\text{GTTP}}$ & P2P+DirInv 
& \makecell{\textbf{27.25} \\ \scriptsize(+0.03)} 
& \makecell{\textbf{0.853} \\ \scriptsize(+0.0054)} 
& \makecell{\textbf{0.05} \\ \scriptsize(-0.0045)} 
& \makecell{25.99 \\ \scriptsize(+0.97)} \\
VENUS w$/_{\text{GTTP}}$ & LEDIT++ 
& \makecell{\underline{23.54} \\ \scriptsize(+0.36)} 
& \makecell{\underline{0.8288} \\ \scriptsize(+0.007)} 
& \makecell{\underline{0.084} \\ \scriptsize(-0.0015)} 
& \makecell{\textbf{26.89} \\ \scriptsize(+0.04)} \\
\bottomrule
\end{tabular}
\label{tab:compareBackbone}
\end{table}

\begin{table}[!t]
\centering
\renewcommand{\arraystretch}{1.1}
\caption{Comparison of different MMLs and the improvements achieved by VENUS in both text-based settings and scene graph-based settings for background preservation and semantic consistency. GTTP denotes the \textit{Ground-Truth Target Prompt} used during editing and evaluation. Best score in \textbf{bold}, second best in \underline{underline}.}
\vspace{-2mm}
\begin{tabular}{l|c|c|ccc|c}
\toprule
\textbf{Method} & \textbf{Multimodal LLM} & \textbf{Para.}& \textbf{PSNR $\uparrow$} & \textbf{SSIM $\uparrow$} & \textbf{LPIPS $\downarrow$} & \makecell[c]{\textbf{CLIP} \\ \textbf{Whole} $\uparrow$} \\
\midrule
\multicolumn{6}{l}{\textit{\textbf{Text-based Editing Approaches}}} \\
VENUS w$/_{\text{GTTP}}$&Qwen2.5-VL-instruct& 7B& \underline{23.54}
& \underline{0.8288} 
& \underline{0.084}
& \underline{26.89} \\
VENUS w$/_{\text{GTTP}}$&Phi-3.5-vision-instruct& 4.15B& \textbf{23.58}
& \textbf{0.8256} 
& \textbf{0.081}
& 26.87 \\
VENUS w$/_{\text{GTTP}}$&InternVL2.5& 4B& 23.24
& 0.8217 
& 0.085
& \textbf{26.92} \\
\midrule
\multicolumn{6}{l}{\textit{\textbf{Scene graph-based Editing Approaches}}} \\
VENUS &Qwen2.5-VL-instruct& 7B& \underline{24.80} & \textbf{0.837} & \textbf{0.070} & \textbf{24.97}\\
VENUS &Phi-3.5-vision-instruct& 4.15B& 24.62 & 0.834 & \underline{0.073} & \underline{24.84}\\
VENUS &InternVL2.5& 4B& \textbf{25.08} & \underline{0.835} & \underline{0.073} & 23.65\\
\bottomrule
\end{tabular}
\label{tab:compareMLLM}
\end{table}

\section{Conclusion}\label{sec:conclusion}

In this work, we introduced VENUS, a training-free approach for image editing guided by scene graphs. By leveraging a split-prompt strategy that disentangles edited content from preserved background, VENUS enables diffusion models to achieve semantic controllability and background fidelity, overcoming a limitation of prior methods. Beyond scene graph-based methods, VENUS also enhances prominent text-based editing backbones, consistently improving performance. 



For future work, we plan to further investigate the potential of lightweight multimodal LLMs, which our experiments suggest can achieve performance comparable to that of larger counterparts. In addition, we aim to explore the use of richer scene graph annotations, including bounding boxes/masks, to enhance position editing accuracy by more precisely delineating true object boundaries.


\textbf{Acknowledgment}. This research is funded by Vietnam National University HoChiMinh City (VNU-HCM) under a project within the framework of the Program titled ``Strengthening the capacity for education and basic scientific research integrated with strategic technologies at VNU-HCM, aiming to achieve advanced standards comparable to regional and global levels during the 2025-2030 period, with a vision toward 2045''

\newpage
\bibliographystyle{splncs04}
\bibliography{main}

\end{document}